\documentclass[runningheads]{llncs}
\usepackage[T1]{fontenc}
\usepackage{graphicx}

\usepackage[switch]{lineno}
\renewcommand{\baselinestretch}{.956}
\usepackage{ulem}
\usepackage{cancel}

\usepackage{ifthen}
\usepackage{fancyhdr}
\usepackage{marvosym}
\usepackage{fontawesome}

\renewcommand{\headrulewidth}{0pt}

\usepackage[utf8]{inputenc} %
\usepackage[T1]{fontenc}    %
\usepackage{url}            %
\usepackage{booktabs}       %
\usepackage{amsfonts}       %
\usepackage{nicefrac}       %
\usepackage{microtype}      %
\usepackage{times}
\usepackage{amsmath,bm}
\usepackage{amssymb}
\usepackage{centernot}
\usepackage{algorithm}
\usepackage{float}
\usepackage{wrapfig}
\usepackage{parskip}
\usepackage{multicol}
\usepackage{enumitem}
\usepackage{verbatim}
\usepackage{algpseudocode}

\usepackage{times,amsmath,epsfig,amssymb,multirow,subfig}
\usepackage{subfig}
\usepackage{lineno}
\usepackage{framed}
\usepackage[framemethod=TikZ]{mdframed}

\def\ie{\textit{i}.\textit{e}., }%
\def\etc{\textit etc.}%
\def\etal{\textit{et al.}}%

\usepackage{mathtools,amssymb}     %
\usepackage{tikz}
\usepackage{xparse}
\NewDocumentCommand{\xleftrightarrows}{ O{}O{} }{%
\mathrel{%
\vcenter{\hbox{%
\begin{tikzpicture}
  \node[minimum width=1cm,minimum height=1ex,anchor=south,align=center] (a){\text{\vphantom{hg}#1}\\[0.5ex] \vphantom{hg}#2};
  \draw[->] ([yshift=0.35ex]a.west) -- ([yshift=0.35ex]a.east);
  \draw[<-] ([yshift=-0.35ex]a.west) -- ([yshift=-0.35ex]a.east);
\end{tikzpicture}
}}%
}%
}

\def\Complex{\mathbb{C}}

\def\th{^{\hbox{\small th}}}

\def\diag{\mathop{\mathrm{diag}}}

\def\vec#1{{\mathbf{#1}}}
\def\mat#1{{\mathbf{#1}}}
\def\ten#1{{\mathbf{\mathcal#1}}}

\def\T{^{\mathrm{T}}}

\def\mode#1{_{\mbox{\tiny\textrm{#1}}}}

\def\time#1{\mbox{\small$#1$}}

\def\measure{\mode 0}
\def\pixels{\mode x}

\makeatletter
\newcommand*{\dottimes}{}%
\DeclareRobustCommand*{\dottimes}{%
  \mathbin{\mathpalette\dottimes@{}}%
}
\newcommand*{\dottimes@scalefactor}{.75}
\newcommand*{\dottimes@widthfactor}{1.15}
\newcommand*{\dottimes@}[2]{%
  \sbox0{$#1\vcenter{}$}%
  \sbox2{$#1\cdot\m@th$}%
  \hbox to \dottimes@widthfactor\wd2{%
    \hfil
    \raise\ht0\hbox{%
      \scalebox{\dottimes@scalefactor}{%
        \lower\ht0\hbox{$#1\bullet\m@th$}%
      }%
    }%
    \hfil
  }%
}
\makeatother

\def\assign{\mathrel{\mathop:}=}

\def\matize#1#2{{\mat#1}\mode{$[#2]$}}

\def\timesT{\times^{\hskip-1pt\hbox{\tiny\textrm{}T}}}
\def\Ttimes{\,\times^{\hskip-11pt\hbox{\tiny\textrm{}T}}}

\def\T{{}^{\hbox{\tiny\textrm{T}}}}

\def\inv{^{-1\lower2pt\hbox{\hskip-2pt\hbox{\tiny${}$}}}}
\def\pinv#1{^{+\lower2pt\hbox{\hskip-2pt\mbox{\tiny ${}#1$}}}}

\usepackage{xcolor}
\usepackage[
pagebackref=false,
breaklinks=true,
letterpaper=true,
colorlinks,
linkbordercolor=blue,
urlbordercolor=blue,
citebordercolor=blue,
bookmarks=false]{hyperref}

\usepackage{scalerel}
\usepackage{tikz}
\usetikzlibrary{svg.path}

\definecolor{orcidlogocol}{HTML}{A6CE39}
\tikzset{
    orcidlogo/.pic={
        \fill[orcidlogocol] svg{M256,128c0,70.7-57.3,128-128,128C57.3,256,0,198.7,0,128C0,57.3,57.3,0,128,0C198.7,0,256,57.3,256,128z};
        \fill[white] svg{M86.3,186.2H70.9V79.1h15.4v48.4V186.2z}
        svg{M108.9,79.1h41.6c39.6,0,57,28.3,57,53.6c0,27.5-21.5,53.6-56.8,53.6h-41.8V79.1z M124.3,172.4h24.5c34.9,0,42.9-26.5,42.9-39.7c0-21.5-13.7-39.7-43.7-39.7h-23.7V172.4z}
        svg{M88.7,56.8c0,5.5-4.5,10.1-10.1,10.1c-5.6,0-10.1-4.6-10.1-10.1c0-5.6,4.5-10.1,10.1-10.1C84.2,46.7,88.7,51.3,88.7,56.8z};
    }
}

\newcommand\orcidicon[1]{\href{https://orcid.org/#1}{\mbox{\scalerel*{
                \begin{tikzpicture}[yscale=-1,transform shape]
                \pic{orcidlogo};
                \end{tikzpicture}
            }{|}}}}

\begin{document}

\title{Causal Deep Learning\vspace{-.285in}
}%

\author{\hspace{+.175in}M. Alex O. Vasilescu
$^{\thanks{To appear 
\href{https://doi.org/10.1007/978-3-031-78189-6_27}{\color{red}doi.org/10.1007/978-3-031-78189-6\_27}. 
}
}$
\hspace{-.125in}
\textsuperscript{\inst{\orcidicon{0000-0001-6581-6930}}}
\href{mailto:maov@cs.ucla.edu}{\hspace{+.0275in}\color{black}\Letter}
\vspace{-.125in}
\institute{IPAM, University of California, Los Angeles CA, USA
\\
Tensor Vision, Los Angeles CA, USA\\
}
\vspace{-.025in}
}

\maketitle

\begin{abstract}
We derive a set of causal deep neural networks whose architectures are a consequence of tensor (multilinear) factor analysis, %
a framework that facilitates 
causal inference. 
Forward causal questions are addressed with a neural network architecture composed of causal capsules and a tensor transformer. Causal capsules compute a set of invariant causal factor
representations, whose interactions are governed by a tensor
transformation. %
Inverse causal questions are addressed with a neural network that implements the multilinear projection algorithm. The architecture reverses the order of operations of a forward neural network and estimates the causes of effects. 
As an alternative to aggressive bottleneck dimension reduction or regularized regression that may camouflage an inherently 
underdetermined inverse problem, we prescribe modeling different aspects of the mechanism of data formation with piecewise tensor models whose multilinear projections %
produce multiple candidate solutions. %
Our forward and inverse questions may be addressed with shallow architectures, but for computationally scalable solutions, we derive a set of deep neural networks by taking advantage of block algebra. %
An interleaved kernel hierarchy results in doubly non-linear tensor factor models.
The causal neural networks that are a consequence of tensor factor analysis are data agnostic, but are illustrated with facial images. %
Sequential, parallel and asynchronous parallel computation strategies are described. %
\vspace{-.185in}
\end{abstract}

\vspace{-.145in}
\section{Introduction}
\label{sec:Introduction}
\vspace{-.065in}
Neural networks %
are being employed increasingly in high-stakes application areas, such as face recognition  \cite{Taigman14,Huang2012,Sun2013,Chen2015,Xiong2016},
and 
 medical diagnosis~\cite{Kermany2018,Madani2018,topol2019}. %
Developing neural networks that offer causal 
explanations 
for correct results or failures is crucial in establishing trustworthy artificial intelligence.

Causal explanations 
specify the causes, the mechanism,
and the conditions for replicating an observed effect
\cite{Lewis1987,Frassen1980,Woodward2005}. %
 Quantitatively, causality is the direct relationship between two events, A and B, where ``A $\xrightarrow[]{\mbox{causes}}$ B'' means ``the effect of A is B'', a measurable and experimentally repeatable phenomena. 
Once %
verified with either experimental or observational studies, 
the statement ``the effect of A is B'' stays true regardless of new discoveries and 
changes in knowledge~\cite{Holland86B}. 
Causal explanations are a factual understanding, whereas interpretations
-- a concept also advocated for the development of trustworthy artificial intelligence~\cite{Lipton2018interpretability} -- are an understanding relative to a specific reference frame, such as a particular point of view or knowledge base. 
As new knowledge emerges, prior interpretations may be deemed to be inaccurate or invalid, which can undermine their reliability and usefulness in the development of trustworthy artificial intelligence. Interpretations are open to reinterpretation. Therefore, explanations should not be confused with interpretations or speculations masquerading as explanations~\cite{Lipton2019scholarship,Lipton2018interpretability}. 

The validity and robustness of causal explanations depend
on causal model specifications in conjunction with the experimental designs used for
acquiring training data~\cite{Rubin08}. 
Generative artificial 
intelligence research %
  that focuses on unsupervised deep neural networks 
    is not well suited for drawing causal conclusions. %
(Determining the causal factors from unsupervised training data %
is an ill-posed inverse problem since different combinations of the same causal factors can result in the same outcome.) %
  Prior unsupervised deep neural network research~\cite{desjardins2012,Bengio2013chapter,Memisevic2010,Tang13} 
  briefly addressed the connection to tensor (multilinear) factor analysis.  
 Tensor (multilinear) factor analysis is
  a supervised learning approach that models the causal mechanisms that generate data~\cite{Vasilescu02,Vasilescu05}, 
   estimates the effects of causes~\cite{Vasilescu04,Vlasic05}, %
as well as the causes of effects, given an estimated forward model and constraints on the solution set~\cite{Vasilescu07a,Vasilescu11}.
Data and its representations -- such as those derived from principal or independent component analysis~\etc -- are factorized into a set of invariant, discriminant, and disentangled causal representations~\cite{Vasilescu09}[(4.15)]\cite{Vasilescu02}. %
These causal representations 
maximize the ratio of inter- to intra-class scatter~\cite{Vasilescu02}. %
Thus, a causal tensor model is not only generative, but discriminative. %
  In fact, the most successful deep neural network approaches 
  are also supervised with generative and/or discriminative properties~\cite{Krizhevsky2012,Taigman14,mathieu2016}.
 However, their goal has been prediction rather than causal inference. %

\begin{figure}
    \centering
     \includegraphics[width=.95\linewidth]{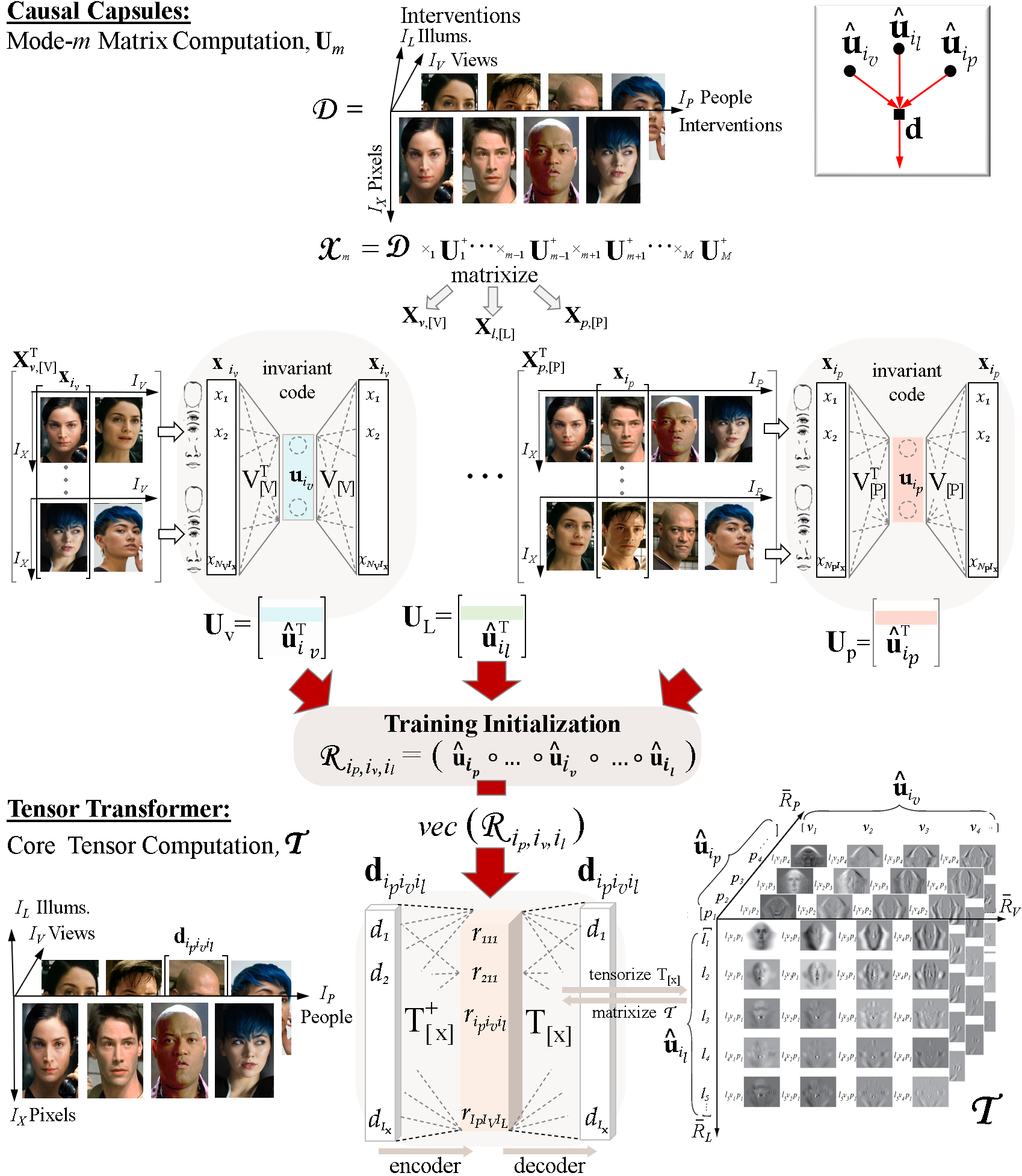}
     \caption{A causal neural network that is an instantiation of the $M$-mode SVD~\cite{Vasilescu02,Vasilescu05}, Alg.~\ref{alg:m-mode-svd}. 
     A causal neural network is composed of a set of causal capsules and a tensor transformer. %
     Given a set of vectorized and centered training observations organized into  a ``data tensor'' $\ten D$, causal capsules learn a set of $M$ subspaces $\{\mat U\mode m|1\le m\le M\}$ that span the $M$ causal factor representations. %
     The tensor transformer estimates the extended core $\ten T$,
     which governs the interaction between causal factors. %
     A forward causal model employs the estimated tensor decoder decoupled from the encoder %
     to generate a new data vector. %
     A new data vector $\vec d\mode{new}$ is the effect of new interventions, %
     $\mbox{vec } \ten R=(\otimes_{m=1}^M \hat{\vec u}_{m\mode{new}})$, %
 where $\otimes$ is the Kronecker product. For scalable computations, each shallow autoencoder-decoder is replaced by a mathematically equivalent deep network, derived with the aid of block algebra, 
 Fig.~\ref{fig:NN_MmodeSVD_ALS}. %
In the upper-right corner, the causal model is summarized 
with %
graphical tensor notation~\cite{Penrose71}, 
which may also be interpreted as
a structural causal model DAG~\cite{Pearl00}. For display purposes only, the images were displayed in grid form and without being mean centered.
          }
    \label{fig:multilinear_autoencoder_decoder}
\vspace{-.25in}
\end{figure}

Consequently, we derive a set of causal deep neural networks that are a consequence of causal tensor factor analysis, Figs.~\ref{fig:multilinear_autoencoder_decoder}-\ref{fig:NN_mprojection}. 
 Tensor factor analysis is a transparent framework for both forward%
~\cite{Vasilescu04,Vasilescu02} %
 and inverse causal inference~\cite{Vasilescu07a,Vasilescu11}.%
\footnote{TensorFaces is a gentle introduction to causal tensor factor analysis~\cite{Vasilescu02,Vasilescu05}.}

Forward causal inference is a hypothesis-driven process, %
as opposed to a data-driven process, 
that models the mechanism of data formation  and estimates the effects of interventions or counterfactuals~\cite{Pearl00,Imbens2015,Spirtes00,Vasilescu20}. %
Inverse causal ``inference'' estimates the causes of effects given an estimated forward model and constraints on the solution set~\cite{Vasilescu11,Gelman13}.  By comparison, conventional statistics and machine learning model data distributions, %
predict a variable co-observed with another, or perform time series forecasting.

\vspace{-.065in}
\subsection{Causal Inference Versus Regression}
\vspace{-.055in}
Neural networks and tensor factorization methods may %
perform causal inference, 
or simply perform regression from which no causal conclusions are drawn. %
For causal inference,
model specifications (\ie problem definition) %
and experimental design for acquiring training data trump algorithmic or neural network design and analysis, %
Fig.~\ref{fig:problem_setup}.

Causal tensor factor analysis was
employed %
 in the analysis and recognition of facial 
identities~\cite{Vasilescu02,Vasilescu20} with sparseness constraints~\cite{Grimes2005,Qiu2015}, as well as in analysis and recognition of facial
expressions~\cite{Wang03b}, human motion signatures~\cite{Davis03,Vasilescu02a}, %
and 3D sound~\cite{Grindlay2007}. %
 It 
  was employed in 
  the rendering of 
arbitrary 
scenes, views and illuminations~\cite{Vasilescu04}, and in the transfer of facial expressions~\cite{Vlasic05}
\etc 
 Tensor factor analysis    
was 
also employed 
 in psychometrics
\cite{Tucker66,Harshman70,Carroll70,Bentler79,Kroonenberg80}, econometrics \cite{Kapteyn86,Magnus88}, 
chemometrics\cite{Bro97}, and signal processing~\cite{Delathauwer00a}. %
Simple tensor regression and decompositions, which do not draw causal conclusions,  
leveraged row, column, and fiber redundancies %
to %
estimate missing data~\cite{chu09}
and to perform rank reduction~\cite{Ye05,Wang08,Bulat20,Iwen21,Hoover21,Benesty21}.
Recently, tensor dimensionality reduction and contractions %
have been employed in machine learning to reduce neural network parameters. Network parameters are organized into ``data tensors'', and dimensionally 
reduced
\cite{Lebedev14,Novikov2015,Kim15,Khrukov20,Onu20} %
or efficiently contracted~\cite{dudek2019}.

\vspace{-.075in}
\subsection{Causal Neural Networks}
\vspace{-.05in}
 Causal neural networks are composed of causal capsules and a  tensor transformer, Fig.~\ref{fig:multilinear_autoencoder_decoder}. 
 {\it Causal capsules} estimate the latent variables that represent the causal factors of data formation, and a {\it tensor transformer}  governs their interaction. %
Causal capsules may be %
shallow %
autoencoder-decoder architectures that employ linear neurons and compute a set of %
invariant representations ~\cite{Sejnowski89,Sanger89,Rumelhart86,Ackley85,Oja82}, as detailed in 
Supplemental {\href{https://openreview.net/pdf?id=AG8yyZQ7-w}{A}. %
The 
tensor transformer 
may be a {\it tensor autoencoder-decoder}, a shallow  autoencoder-decoder whose code is the tensor product of the latent variables. %

\vspace{-.05in}
\begin{figure}[!t]%
\centering
\begin{minipage}{\linewidth}
\centerline{\hfill
\includegraphics[width=.87\linewidth]{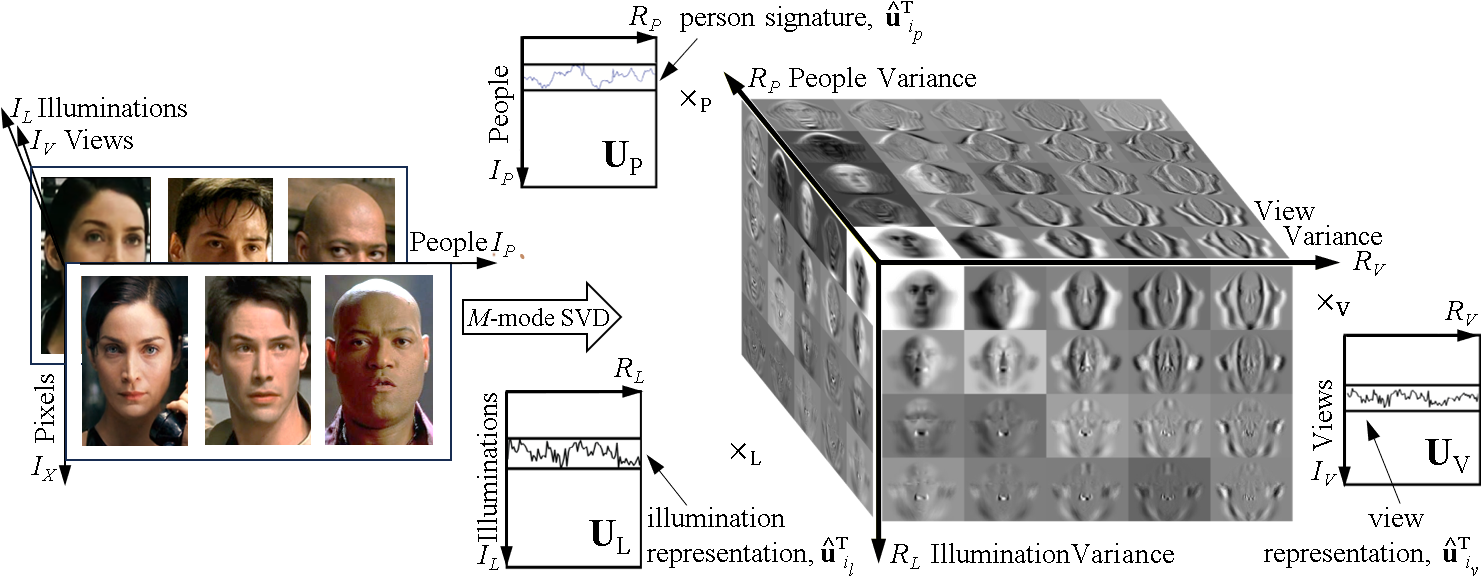}
\hfill}
\centerline{\hfill\mbox{(a)}\hfill}
\centerline{\hfill
\includegraphics[width=.8\linewidth]{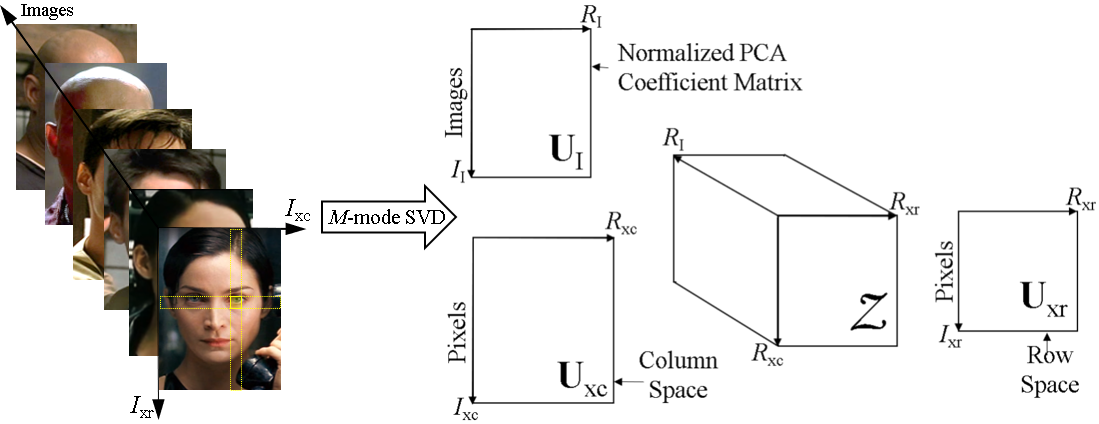}
\hfill
}
\vskip-.05in
\centerline{\hfill\mbox{(b)}\hfill}%
\end{minipage}
\vskip-.1in
\caption{
Same data, same algorithm, but two different model specifications (problem setups / definitions). %
(a) Causal Inference: 
The $M$-mode SVD (Alg.~\ref{alg:m-mode-svd})  
factorizes a ``data tensor'' of vectorized observations into multiple subspaces that span the causal factor representations.
(b) Simple regression: %
The $M$-mode SVD factorizes a ``data tensor'' composed of images as a 2-way array of column or row  observations into a column and row space, as well as into normalized PCA coefficients.
All images are vectorized except in Fig.~\ref{fig:problem_setup}b.}
\label{fig:problem_setup}
\vspace{-.2in}
\end{figure}

Causal deep neural networks are composed by 
    stacking  
    autoencoders-decoders.
Each %
autoencoder-decoder in a shallow causal neural network %
is replaced 
by mathematically equivalent deep neural network architectures that are derived by taking advantage of block algebra. 
An interleaved hierarchy of kernel functions~\cite{Scholkopf98} serves as a pre-processor that warps the data manifold for optimal tensor factor analysis.
A part-based deep neural network %
mirrors a part-based hierarchy of tensor factor models %
\cite{Vasilescu20,Vasilescu19}\cite[Sec 4.4]{Vasilescu09},
Supplemental {\href{https://openreview.net/pdf?id=AG8yyZQ7-w}{C}.%
\footnote{There have been a number of related transformer architectures engineered and empirically tested with success~\cite{fan2021multiscale,wang2021pyramid,Liu2021}.
}%

Inverse causal neural networks implement the multilinear projection algorithm %
to estimate the causes of effects~\cite{Vasilescu07a,Vasilescu11}. %
A neural network that addresses an underdetermined inverse problem is characterized by a wide hidden layer. Dimensionality reduction removes noise and nuisance variables~\cite{Hinton06DimRed,Tishby2015DLBottleneck}, %
 and has the added benefit of reducing the widths of hidden layers. However, aggressive bottleneck dimensionality reduction may camouflage an
inherently ill-posed problem. 
Alternatively or in addition to dimensionality reduction and regularized regression, we
 prescribe modeling different aspects of the data formation process with piecewise tensor (multilinear) models that return a set of candidate solutions%
 ~\cite{Vasilescu02b}. Candidate solutions are gated to yield the most likely solution.
\\\\

\begin{algorithm}[!t]
\begin{minipage}{\linewidth}
{\bf Input} $\ten D\in\Complex^{I_0\times\dots\times
I\mode  M}$, dimensions $R\mode 0, R\mode 1\dots R\mode m \dots R\mode M$\\
1. Initialize $\mat U\mode m \assign \mat I$, %
$0\le m\le M$
\\
2. %
\mbox{}Iterate until convergence

\begin{itemize}
\item[]\hspace{-.125in} 
For $m \assign 0,\dots,M$,

\begin{itemize}
\item[2.1]
$\ten X\assign \ten D \times\mode 0 \mat U\T\mode 0 \times \dots \times\mode {m-1} \mat U\T\mode{m-1}\times\mode{m+1}\mat U\T\mode{m+1}\dots\times\mat U\T \mode M$

\item[2.2]
Set $\mat U_m$ to the $\tilde R_m$ leading left-singular
vectors of the SVD of $\matize X m$ or SVD of $[\matize X m \matize {X\T} m]$.
\footnote{\label{note:cov}%
The
computation of $\mat U\mode m$ in the SVD $\matize X m = \mat
U\mode m\mat\Sigma\mat V\mode m\T$ can be performed efficiently, depending on
which dimension of $\matize X m$ is smaller, by decomposing either
$\matize X m\matize X m\T=\mat U\mode m\mat\Sigma^2\mat U\mode m\T$ (note that
$\mat V\mode m\T= \mat\Sigma^{+}\mat U\mode m\T\matize X m$) or by decomposing
$\matize X m\T\matize X m = \mat V\mode m\mat\Sigma^2\mat V\mode m\T$ and then
computing $\mat U\mode m = 
\matize X m 
\mat V\mode m 
\mat{\Sigma}^{+}$.\vspace{-.0in}}${}^{,}$
\footnote{\label{note:gradient_descent}
For a neural network implementation, the  SVD of $\matize X m$ is replaced with an %
autoencoder  %
that sequentially computes the orthonormal columns of $\mat U\mode m$/$\mat V\mode n$ by performing gradient descent with the learning parameter $\eta$ or stochastic gradient descent~\cite{Bottou1998,Robbins1951}.
In Fig.~\ref{fig:multilinear_autoencoder_decoder}, the autoencoders
compute the columns of $\mat V\mode m$, where $\vec v\mode{m,r}$ is the $r\th$ column and it represents the weights of the $r\th$ neuron. Matrix
$\mat V\mode{m,r}$ contains the first $r$ columns of $\mat V\mode m$.
\vspace{-.05in}
\\
\hspace{-1.75in}
\rotatebox{90}{$\overbrace{\mbox{~~~~~~~~~~~~~~~~~~~~~~~~~~~~~~~~~~~~~}}^{\mbox{~~~~~~~~ %
Autoencoder~~~~~~~~}%
}$\mbox{~~}}{
\vspace{-1.175in}
\item[]
\mbox{~~~~~~~}
For $r\assign 1\dots R\mode m$.\\ 
\mbox{~~~~~~~~~}
Iterate until convergence\\[-3pt]
\vspace{-.15in}
\mbox{~~~~~~~~~~~~~}
\begin{eqnarray}
\hspace{+.5in}
\Delta \vec v\mode {m,r}\time{(t\hspace{-.025in}+\hspace{-.025in} 1)}
\vspace{-.05in}
&=& 
\eta\left(\mat X\mode {[m]} - \mat V\mode {m,r-1} \time{(t)}\mat V\T\mode {m,r-1} \time{(t)}  
\mat X\mode {[m]}\right)\underbrace{\mat X\T\mode{[m]}\vec v\mode{m,r} \time{(t)}}_{\mbox{code}}\nonumber
\vspace{-.075in}\\
\hat{\vec v}\mode{m,r}\time{(t\hspace{-.025in}+\hspace{-.025in} 1)}%
\vspace{-.05in}
&=& 
\frac{\left(\vec v\mode{m,r}\time{(t)}+ \Delta \vec v\mode{m,r}\time{(t\hspace{-.025in}+\hspace{-.025in} 1)}%
\right)}{\|\vec v\mode{m,r}\time{(t)} + \Delta \vec v\mode{m,r}\time{(t\hspace{-.025in}+\hspace{-.025in}1)}%
\|}\nonumber
\end{eqnarray}
}
\vspace{-.1in}
}
\end{itemize}
\end{itemize}

3. Set $\ten Z \assign
\ten D\times_0\mat U\T\mode 0\dots\times\mode m \mat U\T\mode m \dots \times\mode  M \mat U\T\mode  M\assign \ten X \times \mat U\T\mode M$ \footnote{
The columns in $\matize Z 0$ may be computed by initializing the code of an autoencoder to 
$({\vec U}\mode M \dots \otimes {\vec U}\mode m \dots \otimes {\vec U}\mode 0)$,
where $\otimes$ is the Kronecker product. In Fig.~\ref{fig:multilinear_autoencoder_decoder}, the columns of the extended core $\ten T$ are computed by initializing the code of the autoencoder with $(\mat U\mode M \dots \otimes \mat U\mode m \dots \otimes \mat U\mode 1)\T$ for batch training, and $(\hat{\vec u}\T_{i\mode{M}} \dots\otimes \hat{\vec u}\T_{i\mode{m}} \dots \hat{\vec u}\T_{i\mode{1}} )\T$ when training one observation, $\vec d_{i\mode{1},\dots,i\mode{M}}$, at a time. %
\\
\\
\vspace{-.25in}
}
\\
{\bf Output} mode matrices $\mat U_0,\mat U\mode 1,\dots,\mat U\mode  M$ and core
tensor $\ten Z$.
\medskip
\hrule
\vskip-.09in
\caption[$M$-mode Tensor SVD algorithm]{$M$-mode SVD  (parallel computation)\cite{Vasilescu03,Vasilescu02}} %
\label{alg:m-mode-svd}
\vskip-1.95in
\end{minipage}
\vskip-.05in
\end{algorithm}
\vskip-.075in

\vspace{-.2in}
\section{Forward Causal Question: ``What if?''}
\label{sec:forward}
\vskip-.05in
Forward causal inference is a hypothesis-driven process
that addresses the ``what if'' question.%
What if A  is changed by one unit, what is the expected change in B?
Causal hypotheses 
drive both 
the model specification and the 
experimental design for 
acquiring or generating training data. 
\vspace{-.05in}
\subsection{Training Data}
\vspace{-.025in}
For modeling unit level effects of causes, %
the training data is generated by combinatorially varying each causal factor while holding the other factors  
fixed. The best causal evidence comes from randomized experimental studies. %
When %
randomized experiments
for generating training data are unethical or infeasible,  experimental studies may be approximated 
with carefully designed observational studies~\cite{Rubin08},
such as natural experiments~\cite{angrist1990,card1993,Imbens1994} or by employing the concept of transportability to transfer information from a source population where both experimental and observational studies are available to a target population where only observational studies can be conducted~\cite{Pearl14}.%
\footnote{Gebru \etal's ``Datasheets for datasets'' %
may aid in the approximation of experimental studies~\cite{Gebru2021}. %
}

\subsection{Tensor Factor Analysis Model} 
\vspace{-.025in}
Within the tensor mathematical framework,
an $(M+1)$-way array $\ten D\in\Complex^{I\measure \times I\mode1 \dots \times I\mode m \dots \times I\mode M}$ 
is a collection of 
 vectorized 
 and centered observations
$\{\vec d_{i\mode1\dots i\mode m\dots i\mode M} \in\Complex^{I\measure}\}$   
generated by $M$ causal factors. Causal factor $m$, for $m=1\dots M$, takes one of $I_m$ possible values, indexed by $i_m$, where $i\mode m=1\dots I_m$. %
 An observation and an array of observations may be 
 modeled using
 multilinear (tensor) principal component analysis, 
\vspace{-.04in}
\begin{eqnarray}
\vec d_{i\mode{1},\dots,i\mode{M}} 
&=&
\ten T %
\times\mode 1 \vec {\hat u}\T_{i\mode1} 
\dots \times\mode M \vec {\hat u}\T_{i\mode M}
 + \vec e_{i\mode{1},\dots, i\mode{M}}, \\\
\ten D 
&=&
 \ten T \times\mode 1 \mat U\mode 1\dots \times \mat U\mode m \times\mode M \mat U\mode M +\ten E,
\end{eqnarray}
\vspace{-.27in}

\noindent
where the extended core tensor
  $\ten T$ governs the interaction between the latent variables $\{\vec {\hat u}_{i\mode m}\in \Complex^{\tilde{I}\mode m}|1\le m\le M \mbox{ and } \tilde I\mode m\le I\mode m\}$ 
  that represent the causal factors of data formation, and  $\hat{\vec u}\T_{i\mode m}$ 
  is a row in $\mat U\mode m \in \Complex^{I\mode m \times \tilde{I}\mode m}$, Fig.~\ref{fig:problem_setup}a.\footnote{A tensor $\ten T$ is a multilinear mapping from a set of domain vector spaces to a range vector space, $\ten T: \{\Complex^{I\mode 1} \times \dots\times\Complex^{I\mode M}\}\xrightarrow[]{}\Complex^{I\mode 0}$. In a causal tensor framework, the $M$ domain spaces span the causal factor representations and the range vector space spans the observation space. An $M$-way array of vectorized observations, $\ten D$, is sometimes informally referred to as a ``data tensor'', but it is not an actual tensor. It is preferable to vectorize an image and treat it as a single observation rather than as a collection of independent column/row observations%
~\cite[App. A]{Vasilescu09}. For a basic review of tensor algebra, see Supplemental {\href{https://openreview.net/pdf?id=AG8yyZQ7-w}{B}.}
}
The extended core tensor $\ten T=\ten D \times \mode 1\mat U\T\mode 1\dots\times\mode m\mat U\T\mode m\dots\times\mode M\mat U\T\mode M=\ten Z \times\mode 0 \mat U\mode 0$ 
 contains the vectors that span observation subspace.\footnote{In practice, the measurement mode $\mat U\mode 0$,~\ie the PCA basis matrix, and the core tensor $\ten Z$
 are not computed. %
 } 
 The error  $\vec e_{i\mode 1, \dots,i\mode M}\in\Complex^{I\mode 0}$ is a vector in $\ten E\in\Complex^{I\mode 0 \times I\mode 1\dots\times I \mode M}$ and
$\vec e_{i\mode 1, \dots,i\mode M} =\ten Z \times\mode0 \mat E\mode 0 \times\mode1 \vec {\bm \epsilon}\T_{i\mode1}\dots \times\mode M \vec {\bm \epsilon}\T_{i\mode M}$, where 
 $\vec {\bm\epsilon}_{i\mode{m}}$ drawn from a Gaussian distribution 
 $\ten N(\vec 0,\mat \Sigma\mode m )$ associated with each mode $m$ ranging from 0 to M and $\mat E\mode 0=[\dots \vec {\bm \epsilon}_{i\mode 0}\dots \vec {\bm\epsilon}_{I\mode 0}]\T$.
\\
Minimizing the cost function
\vspace{-.125in}
\begin{eqnarray}
    L 
    &=&
    \|\ten D-\ten T\times\mode1 \mat U\mode 1 ... \times\mode m 
    \mat U\mode m... \times\mode M \mat U\mode M\|+\sum_{m=1}^M \lambda\mode m \|\mat U\T\mode m\times\mode 1 \mat U\mode m- \mat I\|
    \vspace{-.1in}%
    \label{eq:cost_fnc}
\end{eqnarray}
\vskip-.1in

is equivalent to maximum likelihood estimation~\cite{Demster77} of the causal factor parameters, 
assuming the data was generated by the model with additive Gaussian noise.
The cost \\function (\ref{eq:cost_fnc}) is minimized by employing $M$  alternating least squares 
optimizations\cite{DeLathauwer00b,Kroonenberg80}
\vspace{-.125in}
\begin{eqnarray}
\hspace{-.25in}
L\mode m %
& = &
\|\ten X\mode m -\ten T\times\mode m \mat U\mode m\| +\lambda\mode m\|\mat U\T\mode m \times\mode1 \mat U\mode m -\mat I\|, 
\label{eq:M_least_squares}
\end{eqnarray}
\vspace{-.35in}%

where
\vspace{-.25in}

\begin{eqnarray}
\hspace{-.25in}
\ten X\mode{m}
& \assign &
{\ten D \times\mode 1 ... \times\mode{m-1} \mat U\T\mode {m-1} \times\mode{m+1} \mat U\T\mode{m+1} ... \times\mode M \mat U\T\mode M}\hspace{+.125in}
\mbox{- parallel computation of all $\ten X\mode m$\hspace{-.4in}}
\label{eq:parallel}\\
& = &
\ten X\mode m\time{(t\hspace{-.025in}-\hspace{-.025in}1)}\times\mode{n} 
\mat U\T\mode{n}\time{(t)}\mat U\mode{n}\time{(t\hspace{-.025in}-\hspace{-.025in}1)}, 
\hspace{+.1in}
\forall\time{n\ne m},\hspace{+.05in}\mbox{- asynchronous parallel computation}%
\label{eq:asynchronous}\\
&=&
(\ten X\mode{m-1} \times\mode{m-1} \mat U\T\mode {m-1}) \times\mode{m} \mat U\mode{m}
\hspace{+.925in}
\mbox{- sequential computation}%
\label{eq:sequential}
\end{eqnarray}
\noindent
\vspace{-.225in}

The $M$-mode SVD~\cite{Vasilescu02} (Alg.~\ref{alg:m-mode-svd}) %
 minimizes the $M$ alternating least squares (\ref{eq:M_least_squares}) in closed form 
 by employing  $M$ different SVDs. 
 The approach is suitable for parallel (\ref{eq:parallel}), asynchronous (\ref{eq:asynchronous}), or sequential (\ref{eq:sequential}) computation.%
The extended core tensor $\ten T$ is computed by multiplying the data tensor with the inverse mode matrices, $\ten T =
\ten D\times\mode 1 \mat U\T\mode 1 \dots \times\mode m \mat U\T\mode m \dots\times\mode M \mat U\T\mode M$, or  more efficiently 
as $\ten T = \ten X \mode m \times \mat U\T\mode m$ where $\ten X\mode m$ is the last updated one.%

\subsection{Kernel Tensor Factor Analysis Model} %

When 
data $\ten D$ is %
a tensor combination %
of non-linear independent causal 
factors, %
kernel multilinear independent component 
analysis (K-MICA)~\cite[Ch 4.4]{Vasilescu09}  %
 employs the ``kernel trick''~\cite{Scholkopf98,vert2004kernelprimer} as a pre-processing step which makes the data suitable for multilinear independent component analysis~\cite{Vasilescu05} (Alg.~\ref{alg:k-mpca-mica}), 
 \begin{eqnarray}
 \ten D &
 =&
 \ten T \times\mode 1 \mat C\mode 1
 \dots \times \mat C\mode m
 \dots\times\mode M \mat C\mode M +\ten E
\hspace{+.05in} \\
\mat C\mode m &
=&
\vec U\mode m\mat W\mode m \inv{} 
 \end{eqnarray}
where $\mat C\mode m$ are the independent components, $\mat W\mode m$ is a rotation matrix computed either with mutual information, negentropy or higher-order
cumulants, and $\ten E=\ten  Z \times\mode 0 \mat E\mode 0\times\mode 1\mat E\mode 1 \dots\times\mode M \mat E\mode M $ is the error. 
K-MPCA is a tensor generalization of the 
kernel PCA~\cite{Scholkopf98} and K-MICA is a tensor generalization of 
kernel 
ICA\cite{Bach2002KICA,Yang05}. %
 
To accomplish this analysis, recall that
the computation %
of
covariance matrix $\matize D m \matize D m\T$ involves inner products
$\vec d\T_{i\mode 1 \dots i\mode{m-1}\, j \,i\mode{m+1}\dots i\mode M} \vec d_{i\mode 2 \dots
i\mode {m-1}\, k \,i\mode {m+1}\dots i\mode M}$ between pairs of data points in the %
data tensor $\ten D$ associated with causal factor $m$, for
$m=1,\dots,M$ (Step 2.2 in Algorithm~\ref{alg:m-mode-svd}). We replace the inner products with a generalized
distance measure between images, $K(\vec d_{i_1\dots
i_{m-1}\,j\,i_{m+1}\dots i\mode M}, \vec d_{i_2\dots
i_{m-1}\,k\,i_{m+1}\dots i\mode M})$, where $K(\cdot,\cdot)$ is a suitable
kernel function (Table~\ref{tab:kernel_functions}) that corresponds
to an inner product in some expanded feature space. This
generalization naturally leads us to a {\it Kernel Multilinear PCA
(K-MPCA) Algorithm},
where the covariance computation %
is replaced by
\vspace{-.05in}
\begin{eqnarray}
\left[\matize D m\matize D m\T\right]_{jk} 
&\assign&
\sum_{i_1=1}^{I_1}\dots
\sum_{i_{m-1}=1}^{I_{m-1}}~\sum_{i_{m+1}=1}^{I_{m+1}}\dots\sum_{i\mode M=1}^{I\mode M}
\nonumber\\
& &%
K(\vec d_{i_1\cdots i_{m-1}\,j\,i_{m+1}\cdots i\mode M}, \vec d_{i_1\cdots
i_{m-1}\,k\,i_{m+1}\cdots i\mode M}).
\label{eq:kernel-covariance-matrix-element}
\end{eqnarray}
When a causal factor is a combination of multiple independent sources that are causal in nature, we employ a rotation matrix $\mat W$ to identify them. The rotation matrix is computed by employing either mutual information, negentropy, or higher-order cumulants~\cite{common94,Bell95,Hyvarinen01,Bartlett02}. A {\it Kernel Multilinear ICA (K-MICA) Algorithm} %
is a kernel generalization of the multilinear independent component analysis (MICA) algorithm  \cite{Vasilescu05}.
 Algorithm~\ref{alg:k-mpca-mica} simultaneously specifies both K-MPCA and K-MICA algorithms. %
A scalable tensor factor analysis represents an observation as a hierarchy of parts and wholes~\cite{Vasilescu20,Vasilescu19}.

\begin{algorithm*}[!t]
\begin{minipage}{\textwidth}
{\bf Input} the data tensor $\ten D\in\Complex^{I_0\times\dots\times I_M}$, where mode $m=0$ is the measurement mode, and the desired ranks are
$\tilde R_1, \dots, \tilde R_M$.

Initialize $\mat C\mode m = \mat I$, $\forall 0\le m\le M$\\
Iterate until convergence.

\begin{enumerate}
\item For $m \assign 1,\dots,M$

\begin{enumerate}
\item Set $\ten X\mode m \assign \ten D\times_1 \mat C_1^+ \dots
\times_{m-1} \mat C_{m-1}^+ \times_{m+1} \mat C_{m+1}^+ \dots
\times_M \mat C_M^+.$%
\item Compute the elements of the mode-$m$ covariance matrix using kernel functions, Table\ref{tab:kernel_functions},
for
 $j,k \assign 1,\dots,I_m$:
\vspace{-.05in}
\begin{equation}
\left[\matize {X\mode m} m\matize {X} m\T\right]\mode{jk}\hskip-.025in \assign\hspace{-.05in}
\sum_{i_1=1}^{I_1}...%
\hskip-.05in\sum_{i\mode{m-1}=1}^{I_{m-1}}~\sum_{i\mode{m+1}=1}^{I\mode{m+1}}%
...\sum_{i\mode{M=1}}^{I_M}
\hskip-,02in K(\vec x_{i \mode 1...%
i\mode{m-1}\,j\,i\mode{m+1}...%
i\mode M}, \vec x_{i\mode 1...%
i\mode{m-1}\,k\,i\mode{m+1}...%
i\mode M}).
\label{eq:kernelized subspace}
\end{equation}
\item\footnote{Every SVD step may be %
autoencoder-decoder. See Algorithm~\ref{alg:m-mode-svd}, Footnotes~\ref{note:cov} and \ref{note:gradient_descent}. \\
See Fig.~\ref{fig:NN_MmodeSVD_ALS} for a scalable neural network implementation.
}
 $\begin{cases}
\text{\sl For K-MPCA:} 
& \text{Set $\mat C\mode m\assign\mat U$, the left matrix of the SVD
of $\left[\matize X m\matize X m\T\right]$ %
from 
(\ref{eq:kernelized subspace})}
\\[-3pt] 
& \text{Truncate to $\tilde R_m$ columns $\mat U\mode m\in\Complex^{I_m\times\tilde R_m}$.}\\[4pt] 
\text{\sl For K-MICA:} 
& \text{Set $\mat C\mode m\assign\mat U\mode m \mat W\inv\mode m$. The additional rotation matrix $\mat W\mode m$ may be}\\[-3pt] 
& \text{computed based on  
negentropy,
mutual information,  or higher-}\\[-3pt] 
    &\text{order cumulants~\cite{Vasilescu05}. The initial SVD of $\left[\matize X m\matize X m\right]\T$ from
    (\ref{eq:kernelized subspace})}\\[-3pt] 
    &\text{truncates the subspace to $\tilde R_m$.}
 \end{cases}$

\end{enumerate}

\item Set $\ten T \assign \ten X\mode M \times_M \mat C_M^+$. For K-MPCA, $\mat C\pinv{} \mode M =\mat C\T \mode M.$

\end{enumerate}

{\bf Output} the converged extended core tensor $\ten
T\in\Complex^{I_0\times\tilde R_1\times\dots\times\tilde R_M}$ and
causal factor mode matrices $\mat C_1, \dots, \mat C_M$.
\end{minipage}
\caption[Kernel Multilinear Independent or Principal Component Analysis (K-MPCA/MICA) algorithm]{Kernel Tensor Factor Analysis 
\cite[Sec 4.4]{Vasilescu09},\cite{Vasilescu05}
\\ 
Kernel Multilinear Independent Component Analysis (K-MICA) and \\Kernel Principal Component Analysis (K-MPCA).}
\label{alg:k-mpca-mica}
\end{algorithm*}
\vskip-.6in
\begin{table*}[!tp]
\vskip-.15in
\centerline{\small
\hskip+.05in
\begin{tabular}{|p{.1in}p{2.35in}p{.25in}p{1.5in}p{.4in}|}\hline
 & & &  & \\[-6pt]
 & Linear kernel: & & $K(\vec u, \vec v)=\vec u\T \vec v=\vec u \cdot \vec v$ &  \\[2pt]
 & Polynomial kernel of degree $d$: & & $K(\vec u, \vec v)=(\vec u\T \vec v)^d$ & \\[2pt]
 & Polynomial kernel up to degree $d$: & & $K(\vec u, \vec v)=(\vec u\T \vec v +1)^d$ & \\[2pt]
 & Sigmoidal kernel: &  & $K(\vec u, \vec v) =\tanh(\alpha\, \vec u\T \vec v + \beta)$ & \\[2pt]
 & Gaussian (radial basis function (RBF)) kernel: & & $K(\vec u, \vec v)=\exp\left(-{\|\vec u - \vec v\|^2 \over 2\sigma^2}\right)$ & \\[7pt]
\hline
\end{tabular}
}
\vspace{+.05in}
\caption[Common kernel functions]{Common kernel functions. Kernel functions are symmetric, positive semi-definite functions corresponding to symmetric, positive semi-definite Gram matrices. The linear kernel does not modify or warp the feature space.}
\label{tab:kernel_functions}
\vskip-.15in
\end{table*}

\newpage 
\subsection{Neural Network Architecture} 
\label{sec:NN}
Tensor factor analysis models are transformed into causal neural networks by using %
autoencoder-decoders as building blocks. 
Causal neural networks 
are composed of causal 
capsules and tensor transformers, Fig.~\ref{fig:multilinear_autoencoder_decoder}. %
{\it Causal capsules} estimate a set of latent variables that represent the causal factors of data formation. %
A {\it tensor transformer} governs the causal factor interaction. %
The M-mode SVD (Algorithm~\ref{alg:m-mode-svd}) is transformed into a neural network by 
replacing every SVD step with gradient descent optimization, which is %
outsourced to a %
autoencoder-decoder with neurons that have a linear transfer function, %
Supplemental {\href{https://openreview.net/pdf?id=AG8yyZQ7-w}{A}.
 For effectiveness, we employ stochastic gradient descent~\cite{Bottou1998,Robbins1951}.
The extended core tensor $\matize T 0$ is computed by defining and employing a tensor autoencoder, an autoencoder whose code is initialized to the Kronecker product, $\otimes$, of the causal factor representations,$\{\vec u_{i\mode m}|1\le i\mode m\le I\mode m \mbox{~~and~~} 1\le m\le M \}$,
\begin{eqnarray}
\vec d_{i\mode 1,\dots,i\mode m \dots i\mode M}&=& \ten T \times\mode 1 \hat{\vec u}\T\mode 1\times \dots\times\mode m \hat{\vec u}\T\mode m \times \dots\times\mode M \hat{\vec u}\T\mode M\\
&=&\matize T 0(\mat  u\T_{i\mode M}\otimes \dots\otimes\vec u\T_{i\mode m} \dots\otimes\vec u\T_{i\mode 1})\T.
\end{eqnarray}
To address a set of arbitrarily non-linear causal factors, each autoencoder employs kernel functions (Algorithm~\ref{alg:k-mpca-mica},Table~\ref{tab:kernel_functions}). %

\subsection{Causal Deep Networks and Scalable Tensor Factor Analysis:}
\begin{figure*}[!t]
\centering
\hskip+1.25in
\includegraphics[width=\linewidth]{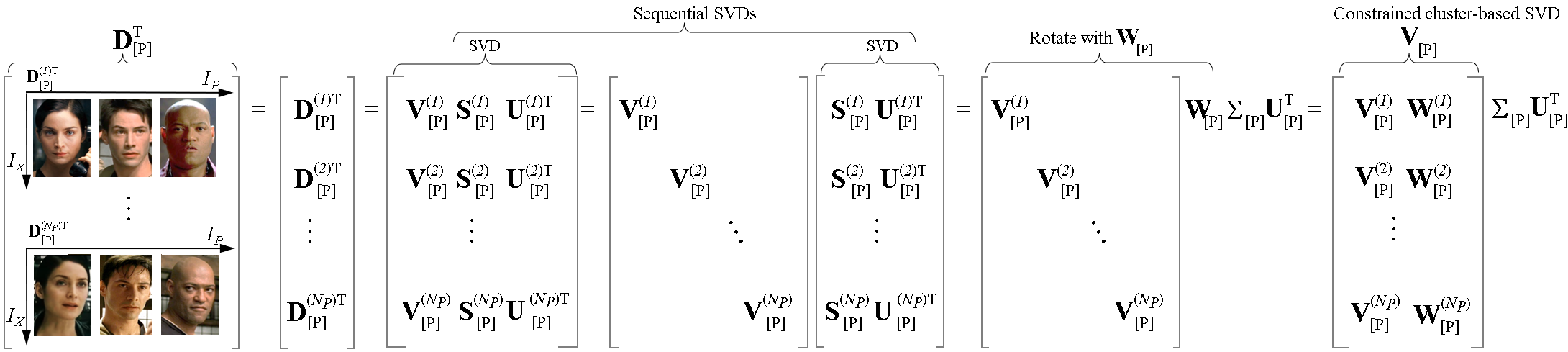}\\
\vskip-.025in
\hskip+.5in\mbox{(a)}\hfill\\
\vspace{-.0in}
\begin{tabular}[b]{c}
\hskip-.025in
\begin{tabular}[c]{ccccccc}
    \includegraphics[height=.19\linewidth]{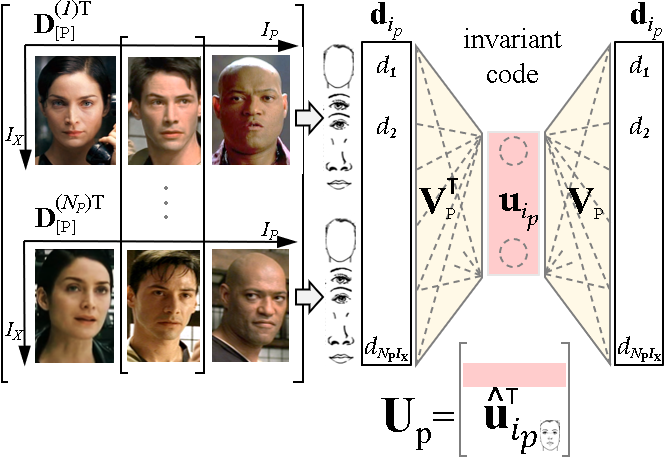}
&
\hskip-.025in
\begin{tabular}{c}
\vspace{+.09in}
\includegraphics[width=.025\linewidth]{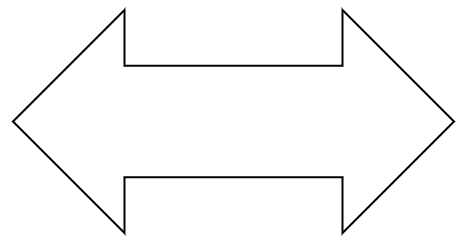}
\vspace{+.9in}
\end{tabular}
&
\hskip-.085in
\includegraphics[height=.188\linewidth]{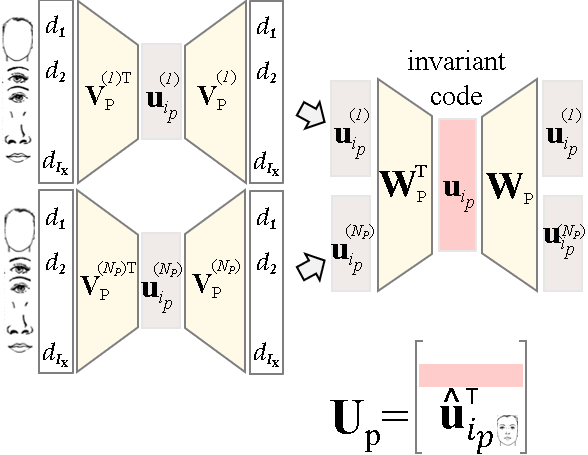}
&
\hskip-.025in
\begin{tabular}{c}
\vspace{+.09in}
\includegraphics[width=.025\linewidth]{images/arrow.png}
\vspace{+.9in}
\end{tabular}
&
\hskip-.085in
\includegraphics[height=.188\linewidth]{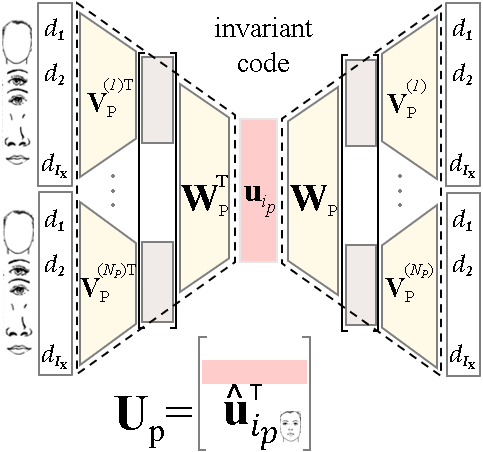}
&
\hskip-.025in
\begin{tabular}{c}
\vspace{+.09in}
    \includegraphics[width=.025\linewidth]{images/arrow.png}
\vspace{+.9in}
\end{tabular}
&
\hskip-.125in
\vspace{+1.4in}
    \includegraphics[height=.188\linewidth]
    {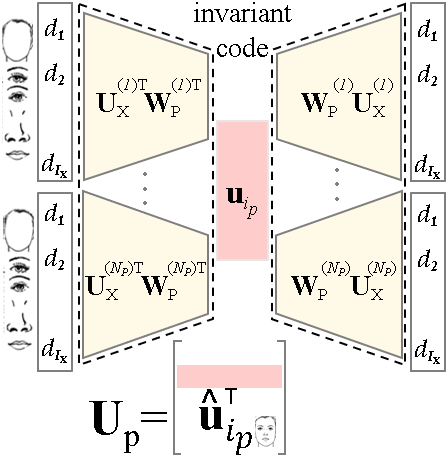}
\vspace{-1.875in}    
\\
\hskip+.675in 
\mbox{(b)}
&&
\hskip+0.625in
    \mbox{(c)}
&&
\hskip
+.02in
    \mbox{(d)}
&&
\hskip-.075in
    \mbox{(e)}
\end{tabular}
\vspace{+.025in}
\\
   \hskip-0.1in
\begin{tabular}[b]{c}
    \includegraphics[width=.7425\textwidth]{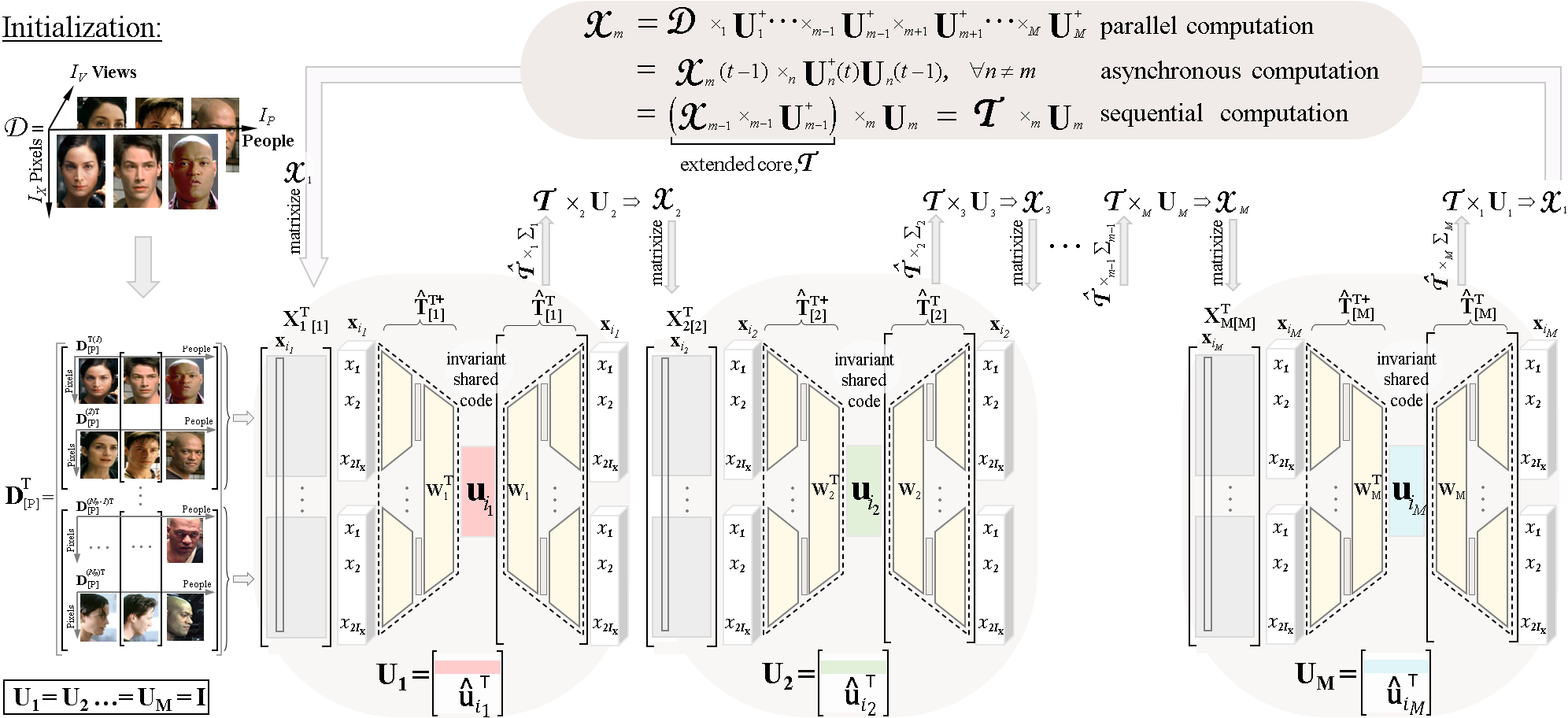}
    \includegraphics[width=.25\textwidth]{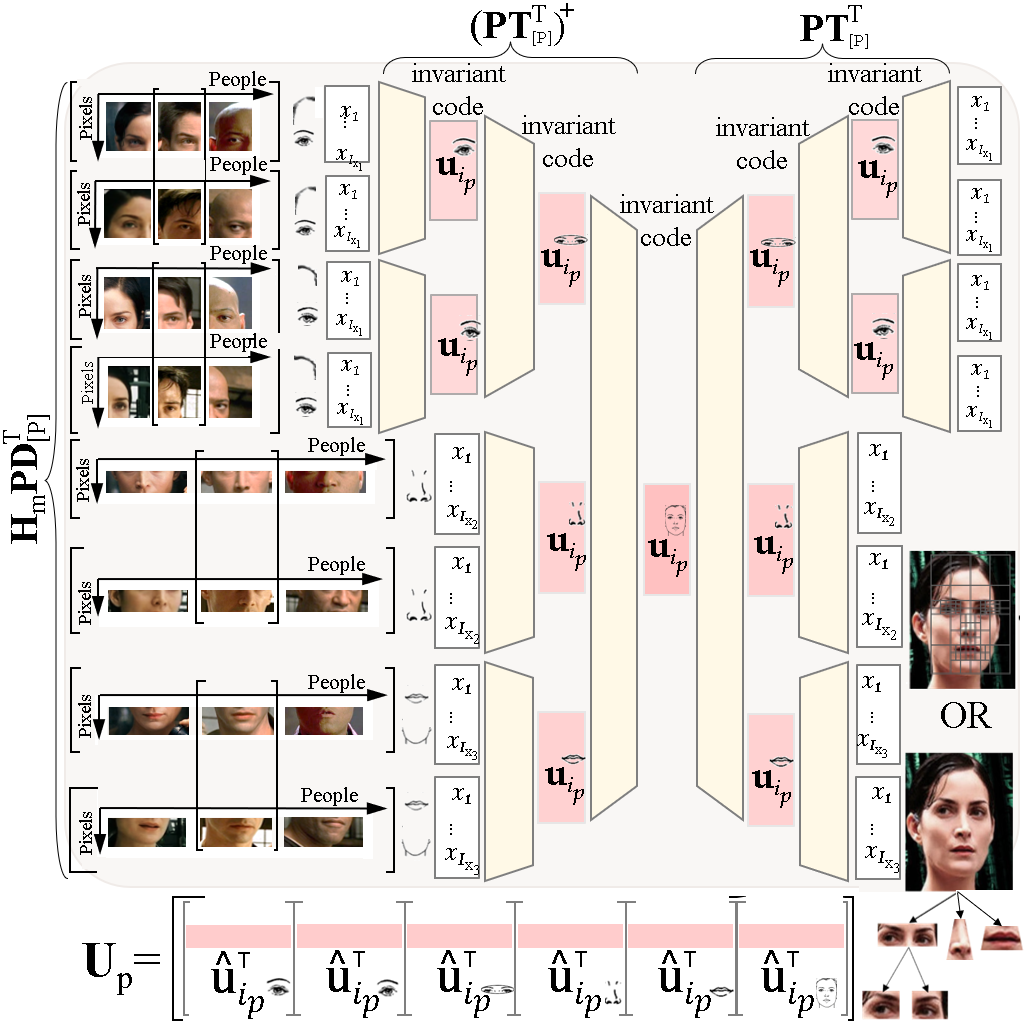}
        \vspace{-.049in}
\\
    \hspace{+1.75in}(f)\hspace{+2in}(g)\hfill
\end{tabular}
\end{tabular}
\vspace{-.2in}
    \caption{Deep neural network. Subfigures (a-e) depict (\ref{eq:block_algebra}--\ref{eq:constrained_cluster_svd3}). 
    (a) The mode matrix computation $\mat U \mode m$ may be thought as a constrained cluster-based PCA that %
    is rewritten in terms of block SVDs. Matrix4izing $\ten D$ may be viewed as a concatenation of ``cluster'' data. %
    The matrix $\mat W$ transforms the basis matrix $\mat V^{(n)}\measure$ 
    such that the causal factor representation $\mat U\mode m$ is the same regardless of cluster membership. 
    In a tensor model, there are $M$ different constrained cluster-based PCAs.
    (b) Mode matrix $\mat U\mode m$ computation using a single autoencoder-decoder. (c) Mode matrix computation as a hierarchy of autoencoder-decoders (\ie feed-forward~\cite{Larochelle07}), (d) Mode matrix computation written as a deep learning model  %
    (e) Concurrent-autoencoders; \ie constrained cluster-based autoencoders. %
       (f) Forward causal model with a set of capsules implemented by deep neural networks. %
       For a parallel, synchronized or asynchronous computation, we break the chain links and shuttle causal information, $\mat U\mode m$, between capsules to compute $\ten X(t+1)$ for the next iteration. %
       (g) Each capsule in (f) may be replaced with a part-based deep neural network by permuting the rows in $\matize {D\T} m$ with $\mat P$, segmented by $\mat H \mode m$, which is efficiently trained with a part-based hierarchy of autoencoders, Fig.~\ref{fig:parts-to_wholes}.}
    \label{fig:NN_MmodeSVD_ALS}   
    \vskip-.25in
\end{figure*}
For a scalable architecture, we  leverage the properties of block algebra. %
Shallow autoencoders are replaced with either a mathematically equivalent deep neural network that is  %
a part-based hierarchy of autoencoders-decoders (\ie feed-forward of Restricted Boltzman Machines~\cite{Larochelle07}), %
 or a set of concurrent autoencoders-decoders, Fig.~\ref{fig:NN_MmodeSVD_ALS}. %

For example, the orthonormal subspace of a data batch, $\mat D\in\Complex^{I\mode 0 \times I\mode 1}$  that has $I\mode 0$ measurements and $I\mode 1$ observations may be computed 
by recursively subdividing the data, analyzing the data blocks, dimensionally reducing their representations, and finally  merging the information %
\vspace{-.075in}
\begin{eqnarray}
\hspace{-.05in}\mat D
&=&
\left[\begin{array}{cc}
    \mat D\mode A\\
    \mat D\mode B 
    \end{array}\right]=
    \left[\begin{array}{c}
    \mat U\mode A \mat S\mode A \mat V\T\mode A\\
    \mat U\mode B \mat S\mode B \mat V\T\mode B 
    \end{array}\right]
=
\left[\begin{array}{cc}
    \mat U\mode A & \mat 0\\
    \mat 0 & \mat U\mode B 
    \end{array}\right] 
    \underbrace{\left[\begin{array}{c}
    \mat S\mode A{\mat V}\T\mode A \\
    \mat S\mode B{\mat V}\T\mode B
    \end{array}\right]}_{\mbox{SVD}}
=
\left[\begin{array}
{cc}
    \mat U\mode A & \mat 0\\
    \mat 0 & \mat U\mode B 
    \end{array}\right] 
\mat W \mat \Sigma \mat V\T
\nonumber\\
\nonumber\\
&=&
\left[\begin{array}
{cc}
    \mat U\mode A & \mat 0\\
    \mat 0 & \mat U\mode B 
    \end{array}\right] 
\left[\begin{array}
{cc}
    \mat W\mode {AA} & \mat W\mode {AB}\\
    \mat W\mode{BA} & \mat W\mode {BB} 
    \end{array}\right] 
\mat \Sigma \mat V\T
=
\left[\begin{array}{c}
    \mat U\mode A \mat W\mode {AA}\\
    \mat U\mode B \mat W\mode {BB} 
    \end{array}\right] 
\mat \Sigma \mat V\T
=\mat U\mat\Sigma \mat V\T.%
\label{eq:block_algebra}
\end{eqnarray}
\vspace{-.2in}

The rotation matrix $\mat W$ transforms basis matrices, $\mat U\mode A$ and $\mat U \mode B$, which span the subspaces of the observations in the data blocks $\mat D\mode A$ 
  and 
$\mat D\mode B$, 
such that the
 observations in $\mat D\mode A$ and the corresponding observations in $\mat D\mode B$ no longer have distinct representations $\mat V\T\mode A$ and $\mat V\T \mode B$, but have the same representations $\mat V\T$. The representations $\mat V\T\mode A$ and $\mat V\T\mode B$ may be dimensionally reduced before moving onto the next step.%
\footnote{Block algebra may be employed if the tensor model is multilinear (tensor) principal component analysis (MPCA), multilinear (tensor) independent component analysis (MICA)\cite{Vasilescu05}, Kernel-MPCA or Kernel-MICA\cite{Vasilescu09}.}

Computing causal factor representations, the mode matrices $\mat U\mode m$ %
of an MPCA tensor model,  %
is equivalent to computing 
a $M$ different of mutually
constrained, cluster-based PCA, 
Fig.~\ref{fig:NN_MmodeSVD_ALS}a. 
When dealing with data that can be 
separated into clusters, the standard 
machine learning approach is to
compute a separate PCA. When data from different clusters are 
generated by the same underlying process (e.g., facial images of the 
same people under different viewing conditions), the  data blocks 
can be concatenated in the measurement mode and the common causal 
factor can be modeled by one PCA. However, for a scalable solution, we employ block algebra (\ref{eq:block_algebra}) and compute a set of constrained cluster-based PCAs, \ie a set of concurrent PCAs. %
Thus, we define a {\it constrained, cluster-based PCA} as the computation of a 
set of PCA basis vectors, such that the latent representation 
is constrained to be the invariant of the cluster membership.

\begin{figure}[!t]
\begin{tabular}{cc}
\includegraphics[width=.58\linewidth]{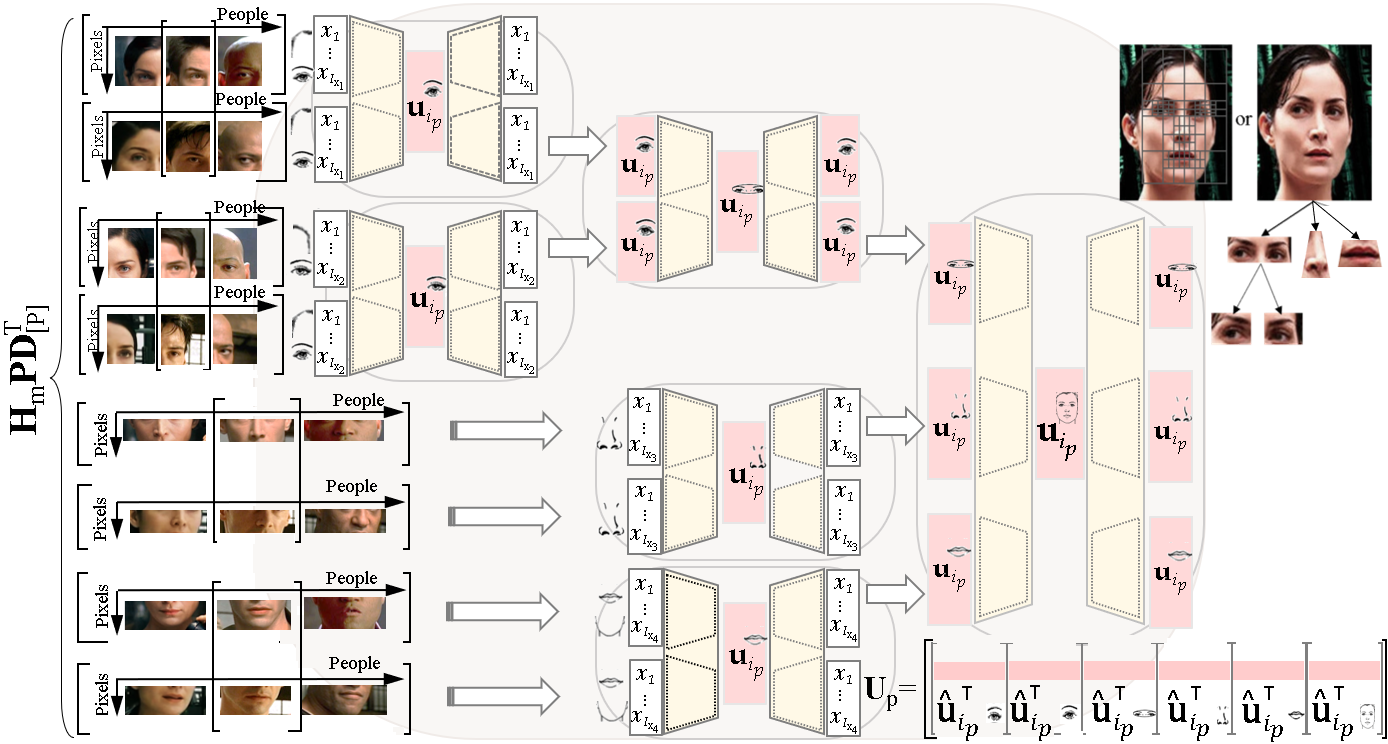}
&
\hspace{+.5in}
\includegraphics[width=.16\linewidth]{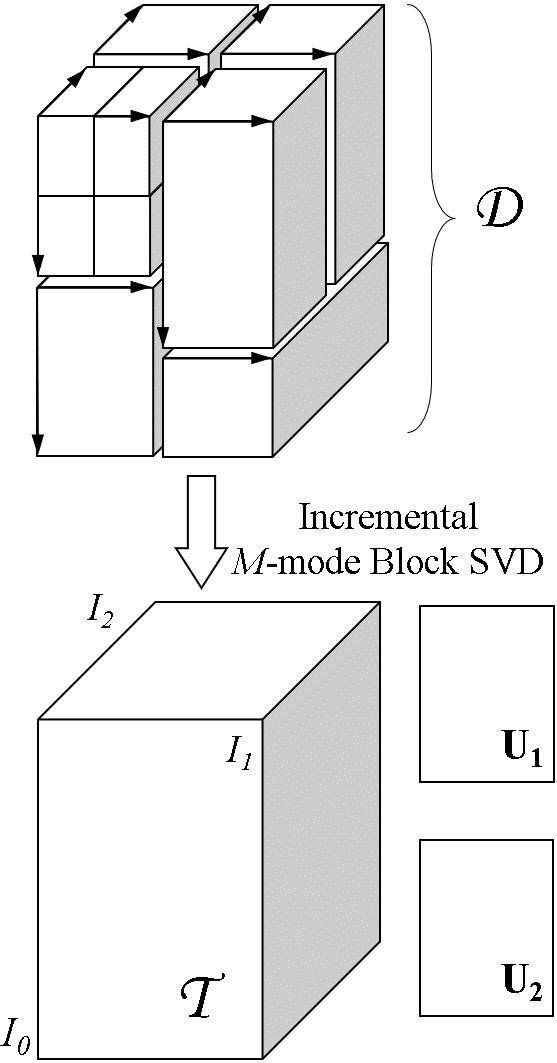}\\
\vspace{+.05in}
\mbox{\hspace{+.25in}(a)}&\hspace{+.5in}\mbox{(b)}
\end{tabular}
\vspace{-.175in}
\caption{(a) Causal capsules may be implemented with a part-based hierarchy of autoencoders. The dataset is permuted by $\mat P$, segmented and filtered by $\mat H\mode m$ that is mode dependent. %
(b) Implementing the capsules with a part-based hierarchy of autoencoders is equivalent to performing M-mode Block SVD\cite[Sec IV]{Vasilescu20}\cite{Vasilescu06b}.}
\label{fig:parts-to_wholes}
\vspace{-.15in}
\end{figure}
In the context of our multifactor
data analysis, we define a cluster as a set of observations for which
all factors are fixed but one.
For every tensor mode, there
are 
$N\mode m=I_1I_2 \dots I_{m-1}I_{m+1} \dots I\mode  M$ possible clusters and
the data in each cluster varies with the same causal
mode. 
The constrained, cluster-based PCA concatenates the clusters in the
measurement mode and analyzes the data with a linear model, such as
PCA.%

To see this, let $\ten D_{i_1\dots i_{m-1}i_{m+1}\dots
i\mode  M}\in\Complex^{I_0\times1\times1\dots\times1\times
I\mode m\times1\dots\times1}$ denote a subtensor of $\ten D$ that is
obtained by fixing all causal factor modes but mode $m$ and mode 0
(the measurement mode). Matrixizing this subtensor in the measurement mode we
obtain $\matize {D_{i_1\dots i_{m-1}i_{m+1}\dots i\mode  M}}
0\in\Complex^{I_0\times I\mode m}$. This data matrix comprises a cluster of
data obtained by varying causal factor $m$, to which one can
traditionally apply PCA. Since there are $N\mode m=I_1I_2 \dots
I_{m-1}I_{m+1} \dots I\mode  M$ possible clusters that share the same
underlying space associated with factor $m$, the data can be
concatenated and PCA performed in order to extract the same
representation for factor $m$ regardless of the cluster. Now,
consider the MPCA computation of mode matrix $\mat U\mode m$, Fig.~\ref{fig:NN_MmodeSVD_ALS}a,
which can be written in terms of matrixized subtensors as
\vspace{-.2in}
\begin{equation}
{\mat D\mode{m}} 
= \left[\begin{array}{c} 
{\matize {D_{1\dots 11\dots 1}} m}\T\\
{\tiny\vdots}\\ \matize {D_{I_1\dots 11\dots 1}} m\T\\ {\tiny\vdots}\\
{\matize {D_{I_1\dots I_{m-1}I_{m+1}\dots I\mode  M}} m}\T
\end{array}
\right]^{\T}
= \mat U\mode m \mat\Sigma\mode m \mat V\mode m\T.
\vspace{-.08in}
\end{equation}
This is equivalent to computing a set of $N\mode m=I_1I_2 \dots
I_{m-1}I_{m+1} \dots I\mode  M$ cluster-based 
PCAs concurrently by
combining
them into a single statistical model and representing the underlying
causal factor $m$ common to the clusters. Thus, rather than computing
a separate linear PCA model for each cluster, MPCA concatenates the
clusters into a single statistical model and computes a representation
(coefficient vector) for mode $m$ that is invariant relative to the
other causal factor modes $1,..., (m-1),(m+1),...,M$. For a scalable solution, we rotate the cluster-based PCA basis vectors, such that the data blocks have the same representation regardless of cluster memebership. Thus,
MPCA performs  multiple constrained, cluster-based PCA.
To clarify the relationship, let us number each of the matrices
 $  \matize {D_{i_1\dots i_{m-1}i_{m+1}\dots i\mode  M}} m = \mat D\mode m^{(n)}$
with a parenthetical superscript $1\le n=1+\sum_{k=1, k\neq m}^M(i_n -
1)\prod_{l=1, l\neq m}^{k-1} I_l \le N\mode m$. 
Let each of the cluster SVDs be $\mat D\mode m^{(n)} 
=\mat U\mode m^{(n)}\mat\Sigma\mode m^{(n)}{\mat V_m^{(n)}}\T$, and 
\begin{eqnarray}
\matize D m 
&=& 
\underbrace{\left[%
\begin{array}{c}
\mat U\mode m^{(1)}\mat\Sigma\mode m^{(1)} \dots \mat U\mode m^{(N\mode M)}\mat\Sigma\mode m^{(N\mode m)}
\end{array}%
\right]}_{\mbox{SVD}} %
\diag([
\begin{array}{c}
\mat V^{(1)}\mode m  \dots \mat V^{(N\mode m)}\mode m
\end{array}
])\T 
\label{eq:constrained_cluster_svd1}\\
&=& 
\mat U\mode m\mat\Sigma\mode m \mat W\T\mode m
\diag([
\begin{array}{ccc}
\mat V\mode m^{(1)} & \dots & \mat V\mode m^{(N\mode m)}
\end{array}
])\T, 
\label{eq:constrained_cluster_svd2} \\
&=&\mat U\mode m\mat\Sigma\mode m
[
\begin{array}{ccc}
\mat V^{(1)}\mode m \mat W^{(1)}\mode m 
&  \dots  & 
\mat V^{(N\mode m)}\mode m \mat W^{(N\mode m)}\mode m 
\end{array}
]\T %
\\
&=&\mat U\mode m\mat\Sigma\mode m \mat V\mode m\T,
\label{eq:constrained_cluster_svd3}
\end{eqnarray}
\vspace{-.35in}

where $\diag(\cdot)$ denotes a diagonal
matrix whose elements are each of the elements of its vector argument.
The mode matrix $\mat V^{(n\mode m)}\mode m$ is the measurement matrix $\mat U^{(n\mode m)}\measure$ ($\mat U^{(n\mode m)}\pixels$ when the measurements are image pixels) that contains the eigenvectors
spanning the observed data in cluster $n\mode m$, $1\le n\mode m\le N\mode m$. 
 MPCA can be thought as computing a rotation matrix, $\mat W\mode m$, that contains  a set of blocks
$\mat W^{(n)}\mode m$ along the diagonal  
 that transform the PCA cluster eigenvectors $\mat V^{(n\mode m)}\mode m$ %
 such that the mode matrix $\mat U\mode m$ is the same regardless of cluster membership (\ref{eq:constrained_cluster_svd1}--\ref{eq:constrained_cluster_svd3}), Fig~\ref{fig:NN_MmodeSVD_ALS}. 
 The constrained ``cluster''-based PCAs may also be implemented with a set of concurrent ``cluster''-based PCAs, Fig.~\ref{fig:NN_MmodeSVD_ALS}e.

Causal factors of object wholes may be computed efficiently from their parts, %
by applying a permutation matrix $\mat P$ and creating part-based data clusters with a segmentation filter $\mat H\mode m$, where %
$\ten D\hspace{+.05in} \Ttimes\mode m \mat H\mode m \mat P \Leftrightarrow  \mat H\mode m\mat P {\matize D m}\T$. %
 A deep neural network can be efficiently trained with a hierarchy of part-based autoencoders, Fig.~\ref{fig:parts-to_wholes}. %
A computation that employs a part-based hierarchy of autoencoders parallels the Incremental M-mode Block SVD~\cite[Sec. IV]{Vasilescu20,Vasilescu06b}.
A data tensor is recursively subdivided into data blocks, analyzed in a bottom-up fashion, and the results merged as one moves through the hierarchy. %
The computational cost is the cost of training one autoencoder, $\mathcal{O}(T)$, times $\mathcal{O}(log N\mode M)$, the total number of autoencoders trained for each factor matrix, $\mathcal{O}(T log N\mode m)$. If the causal neural network is trained sequentially, the training cost for one time iteration is $\mathcal{O}(M T log \bar N)$, where $\bar N$ is the average number of clusters across the $M$ modes.

\vspace{-.075in}
\section{Inverse Causal Question: ``Why?''}
\label{sec:inverse}
\vspace{-.075in}
Inverse causal inference addresses the “why” question and 
estimates the causes of effects 
given an estimated forward causal model and a set of constraints that reduce the solution set\footnote{As discussed earlier in the paper, different combinations of the same causal factors can lead to the same outcome. In imaging, this phenomena may result in visual illusions. This is a many-to-one problem,
and its inverse is ill-posed without constraints. 
} and render the problem well-posed~\cite{Gelman13,Vasilescu11,Vasilescu07a}.

Multilinear tensor factor analysis
constrains causal factor representations to be unitary vectors. Multilinear projection~\cite{Vasilescu07a,Vasilescu11} relies on this constraint and performs multiple regularized regressions. One or more unlabeled test observations %
that are not part of the training dataset are multilinearly projected 
into the causal factor spaces, %
\vspace{-.185in}

\begin{equation}
\begin{array}{ccl}
\ten T\pinv{0}\timesT\mode 0 {\vec d}\mode{test} 
&=&
\ten R%
\\
&=&\mbox{CP}(\ten R) \mbox{~~or~~ M-mode SVD}(\ten R)\\
&\approx& 
\vec{\hat{r}}\mode 1
... \circ \vec{\hat{r}}\mode m ... \circ\vec{\hat{r}}\mode M, %
\end{array}
\end{equation}
\vspace{-.175in}

\noindent
where $\vec{\hat{r}}\mode m$ is the estimated orthonormal causal factor $m$ representation, $\circ$ is the outer product and $\ten T\pinv{0}$ is the tensorized $\matize T 0\pinv{}$. 
A %
neural network that implements a multilinear projection architecture 
is an inverted (upside-down) forward neural network architecture that employs an estimated $\matize T X\pinv{}$ and reverses the operation order, Fig.~\ref{fig:NN_mprojection}. 

Neural architectures addressing underdetermined inverse problems are characterized by hidden layers that are wider than the input layer; \ie the dimensionality of $vec(\ten R)$ is larger than the number of measurements in $\vec d$. Dimensionality reduction reduces noise, and the width of the hidden layers~\cite{Hinton06DimRed}. However, it can also camouflage an inherently underdetermined inverse problem. %
Adding sparsity, non-negativity constraints\cite{Vendrow2021}, etc., can further reduce the solution set in a principled way.
Alternatively or in addition, one can determine a set of candidate solutions 
by modeling different aspects of the mechanism of data formation as piecewise tensor (multilinear) factor  models. %
A single multilinear projection~\cite{Vasilescu07a,Vasilescu11} is replaced with multiple multilinear projections. %
Vasilescu and Terzopoulos~\cite{Vasilescu02b} rewrote the forward multilinear model in terms of multiple piecewise linear models  %
that were employed to perform multiple linear projections and produced multiple candidate solutions that were gated to return the most likely solution.
\begin{figure}[!t]
\begin{tabular}{p{.44\linewidth}p{.44\linewidth}}
\vspace{-1.955in}
\caption{An inverse causal network %
is an inverted (upside-down) forward network that 
performs its operations in reverse order using %
$\mat T\mode {\mbox{[X]}}\pinv{}$, estimated during training, Fig~\ref{fig:multilinear_autoencoder_decoder}. The architecture implements the multilinear projection\cite{Vasilescu11,Vasilescu07a}. 
For a scalable solution, each autoencoder-decoder is replaced with a deep network.
}
\label{fig:NN_mprojection}
&
\hspace{+.25in}
\includegraphics[width=.99\linewidth]{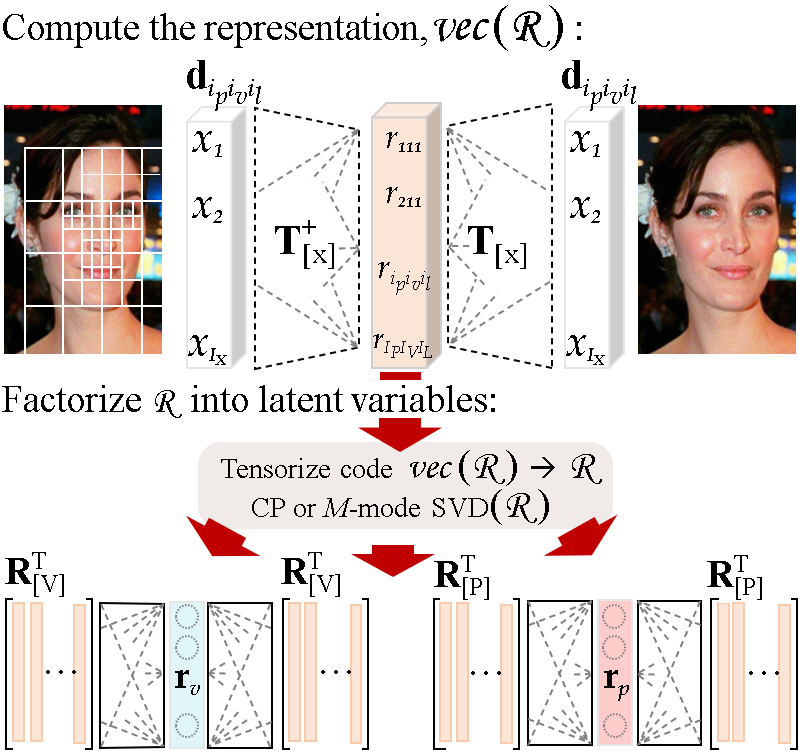}
\end{tabular}
\vspace{-.35in}
\end{figure}

\vspace{-.05in}
\section{Conclusion}
\label{sec:conclusion}
\vspace{-.05in}
We derived a set of shallow and deep causal neural networks that %
are a consequence of causal tensor factor analysis. %
Causal neural networks 
are composed of causal capsules and a tensor transformer. 
Causal capsules compute invariant causal factor representations, whose interaction are governed by a tensor transformation. 
An inverse causal neural network implements the multilinear projection and estimates the causes of effects. %
As an alternative to aggressive ``bottleneck'' dimensionality reduction that may camouflage an inherently underdetermined inverse problem, the mechanism of data formation is modeled as piecewise tensor (multilinear) models, and inverse causal neural networks perform multiple multilinear projections that result in multiple candidate solutions, which may be gated to yield the most likely solution. %

{\bf Acknowledgement:} Demetri Terzopoulos and Ernest Davis provided invaluable feedback on various drafts of this paper.}
\newpage
\clearpage
\appendix
\section*{Supplemental}
\label{sec:math}
\subsection*{A. \href{https://openreview.net/pdf?id=AG8yyZQ7-w}{\color{red}PCA computation with a Hebb Autoencoder}}
\label{sec:autoencoder}
\subsection*{B. \href{https://openreview.net/pdf?id=AG8yyZQ7-w}{\color{red}Relevant Tensor Algebra}}
\label{sec:tensor_review}
\subsection*{C. \href{https://openreview.net/pdf?id=AG8yyZQ7-w}{\color{red}Compositional Hierarchical Block TensorFaces}}
\label{sec:Compositional-TensorFaces}
\newpage
\vspace{-.25in}
{\small
\bibliographystyle{ieee}
\bibliography{all,ekrefs}
}

\end{document}